\pdfoutput=1

\documentclass[11pt]{article}

\usepackage{amsmath,amsfonts,bm}

\def\eqref#1{equation~\ref{#1}}

\def\1{\bm{1}}

\def\ve{{\bm{e}}}

\def\vh{{\bm{h}}}

\def\vp{{\bm{p}}}

\def\vs{{\bm{s}}}

\def\vw{{\bm{w}}}

\def\vy{{\bm{y}}}
\def\vz{{\bm{z}}}

\def\mW{{\bm{W}}}

\def\mZ{{\bm{Z}}}

\DeclareMathAlphabet{\mathsfit}{\encodingdefault}{\sfdefault}{m}{sl}
\SetMathAlphabet{\mathsfit}{bold}{\encodingdefault}{\sfdefault}{bx}{n}

\usepackage[preprint]{acl}
\usepackage{graphicx}
\usepackage{times}
\usepackage{latexsym}

\usepackage{amsmath,amsfonts,bm}
\usepackage[T1]{fontenc}
\usepackage{booktabs}
\usepackage{multirow}
\usepackage{comment}

\usepackage[utf8]{inputenc}

\usepackage{microtype}

\usepackage{inconsolata}

\title{\textit{GraphER}: A Structure-aware Text-to-Graph Model for Entity and Relation Extraction}

\author{Urchade Zaratiana$^{1,2}$, Nadi Tomeh$^2$, Niama El Khbir$^{2}$, Pierre Holat$^{1,2}$, Thierry Charnois$^2$ \\
$^1$ FI Group, 
$^2$ LIPN, CNRS UMR 7030, France \\{\tt zaratiana@lipn.fr} \\
Code: \texttt{\textbf{https://github.com/urchade/GraphER}}}

\begin{document}
\maketitle
\begin{abstract}
Information extraction (IE) is an important task in Natural Language Processing (NLP), involving the extraction of named entities and their relationships from unstructured text. In this paper, we propose a novel approach to this task by formulating it as graph structure learning (GSL). By formulating IE as GSL, we enhance the model's ability to dynamically refine and optimize the graph structure during the extraction process. This formulation allows for better interaction and structure-informed decisions for entity and relation prediction, in contrast to previous models that have separate or untied predictions for these tasks. When compared against state-of-the-art baselines on joint entity and relation extraction benchmarks, our model, \textit{GraphER}, achieves competitive results.
\end{abstract}

\section{Introduction}

Information extraction is a fundamental task in NLP with many crucial real-world applications, such as knowledge graph construction. Early systems for this task were rule-based with manually coded rules \citep{Appelt1993FASTUSAF,10.5555/1867270.1867391}, which are time-consuming and offer low performance. Methods based on machine learning have been proposed \citep{zelenko-etal-2002-kernel,jiang-zhai-2007-systematic}, usually implementing pipeline approaches, with entity and relation models trained separately \citep{roth-yih-2004-linear,rosenfeld-feldman-2007-using}. The emergence of deep learning has enabled the training of joint IE models end-to-end through multitask learning, benefiting from rich features learned from self-supervised language models \citep{peters-etal-2018-deep,Devlin2019BERTPO}.

\paragraph{} Span-based approaches were proposed \citep{dixit-al-onaizan-2019-span,Eberts2019SpanbasedJE,ji-etal-2020-span}, which first classify spans as entities and then predict the relations by classifying all pairs of span entities. While this approach benefits from rich span representations computed by pretrained transformers, it overlooks potential interactions between entities and relations, as relations cannot influence entity types since entities are predicted first. As an alternative, table-filling approaches \citep{gupta-etal-2016-table,wang-lu-2020-two,ren-etal-2021-novel,yan-etal-2023-utc} have been proposed to perform joint predictions, using unified labels for the task, unlike span-based approaches that assume a prediction order. While it obtains competitive performance, by not explicitly modeling the graph nature of the task, table-filling methods might miss out on capturing more complex structural dependencies. More recently, autoregressive approaches have gained popularity, treating this task as linearised graph generation \citep{paolini2021structured}. Several approaches have been proposed, either by fine-tuning \citep{lu-etal-2022-unified,liu-etal-2022-autoregressive,fei2022lasuie} or by prompting large language models in zero or few-shot \citep{wadhwa-etal-2023-revisiting,geng-etal-2023-grammar,Han2023IsIE}. However, these models exhibit slow inference due to autoregressive generation and are prone to issues such as hallucination, where the model generates plausible but incorrect or irrelevant information \citep{guerreiro-etal-2023-optimal,manakul-etal-2023-selfcheckgpt}.

\begin{figure*}[h]
    \centering
\includegraphics[width=1.\textwidth]{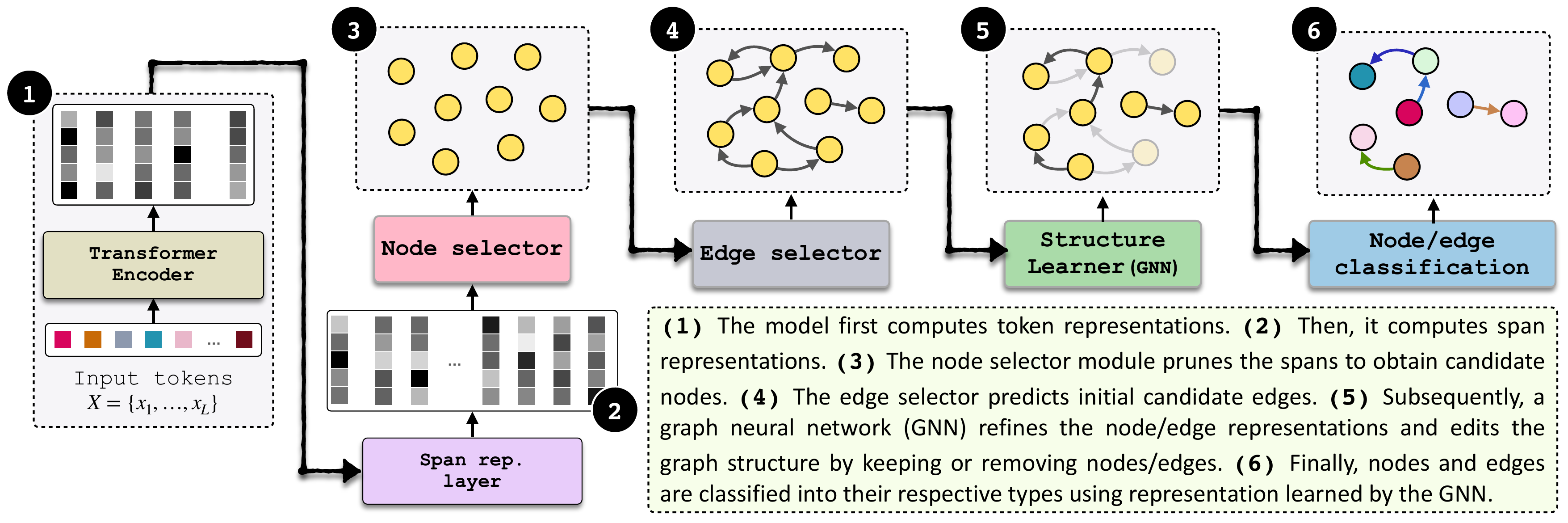}
    \caption{\textit{GraphER} architecture. Please refer to the text box in the figure for an explanation of the different steps.}
    \label{fig:architecture}
\end{figure*}

\paragraph{} In this paper, we propose a new paradigm for the IE task by treating it as a graph structure learning problem \citep{kipf2018neural,franceschi2019learning,jin2020graph,Zhao_Wang_Shi_Hu_Song_Ye_2021,li2023gslb}, enabling more robust graph representation and thus structure-informed prediction. Our model, \textit{GraphER}, begins by creating an initial, imperfect graph from the text, where nodes represent textual spans and edges represent the relationships between these spans. Subsequently, the structure learner performs two operations: 1) it first enriches the representation of the elements of the graph using Graph Neural Networks (GNN) \citep{Kipf2016SemiSupervisedCW,Hamilton2017InductiveRL}, and then 2) it performs edit operations on the current graph, by either \textit{keeping} or \textit{dropping} elements (node or edge) to recover the final graph structure. Our structure learner is similar to the Graph Edit Network \cite{paassen2021graph}, except that our model does not have an \textit{adding} operation as we assume that the initial graph contains them, and all edit operations are performed in a single step in our case. Furthermore, our structure learner takes advantage of recent advancements in GNN literature by employing the Token Graph Transformer (TokenGT) \citep{kim2022pure}, a highly expressive model for graph-structured tasks. We found that it performs significantly better than standard message-passing GNNs \citep{pmlr-v70-gilmer17a} such as graph convolution network (GCN) \citep{Kipf2016SemiSupervisedCW} and graph attention network (GAT) \citep{veličković2018graph}, especially since our graph is noisy and highly heterogeneous (i.e., contains many types of nodes and edges). Finally, our model performs node and edge classification on the final graph structure. When evaluated on benchmark datasets for joint IE, we found that our model achieves competitive results compared to strong baselines.

\section{Input Graph Modeling}
In this section, we provide a detailed explanation of the architecture of \textit{GraphER}, as illustrated in Figure \ref{fig:architecture}.

\subsection{Span Representation}
The first step of \textit{GraphER} consists of converting the input token $\{x_i\}_{i=1}^{L}$ into a set of contextualised embeddings $\{\vh_i\}_{i=1}^{L} \in \mathbb{R}^D$. In this work, token embeddings are computed using a pretrained transformer encoder \citep{Devlin2019BERTPO}. Then, the representation of a span starting at position $i$ and ending at position $j$ is computed as follow:

\begin{equation}
\label{eq1}
    \vs_{ij} = \text{FFN}([\vh_i^s ; \vh_j^e])
\end{equation}

where, $\vh_i^s$ and $\vh_j^e$ are the embeddings of the start and end word respectively, and $\text{FFN}$ is a two-layer feed-forward network. To prevent quadratic complexity, we restrict the maximum width of the spans to a fixed number ($<L$).

\begin{figure*}[h]
    \centering
\includegraphics[width=1.\textwidth]{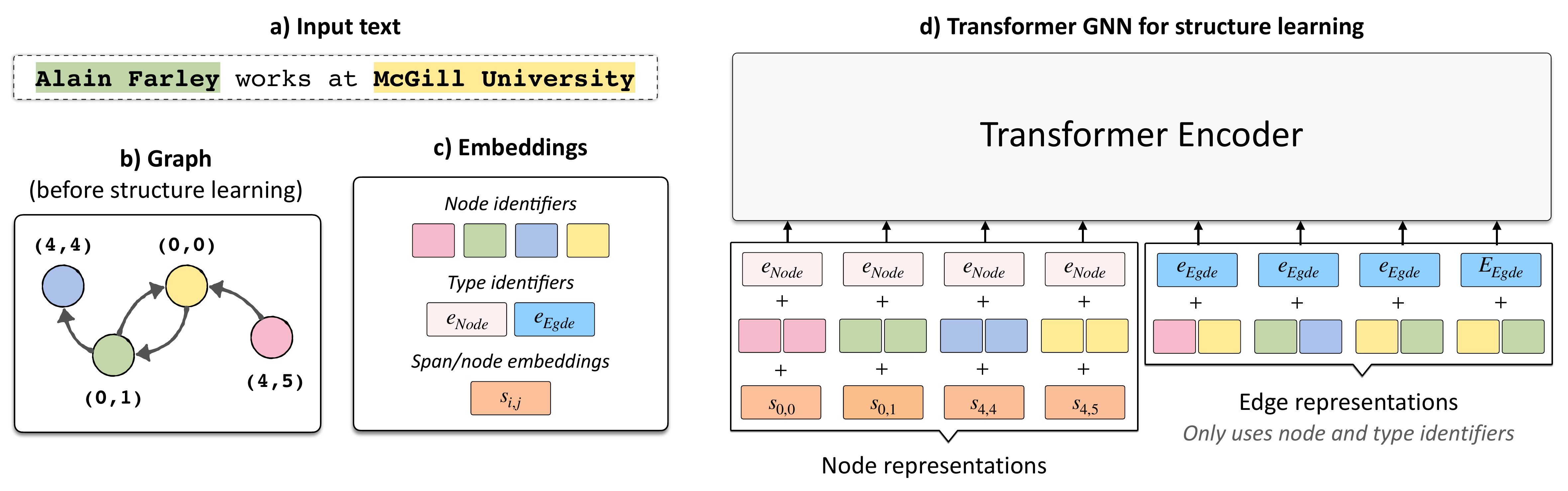}
        \caption{Graph transformer: \textit{GraphER} first constructs an initial graph from the input text (b) (Section \ref{graph_con}). A transformer then processes nodes and edges of the graphs to refine their representation (d) (Section \ref{trans_tok}). Edge representation only uses node and type identifiers (without using edge-specific features) to enforce the transformer to use the graph structure for representation computation.}
    \label{fig:architecture}
\end{figure*}

\subsection{Graph Construction \label{graph_con}}
Here, we describe the initial graph construction process in \textit{GraphER}, which corresponds to steps 2 and 3 of Figure \ref{fig:architecture}. The objective is to select an appropriate number of nodes and edges before the structure learning phase. Including an excessive number of nodes and edges can hinder scalability, while opting for too few may lead to recall issues, potentially omitting critical nodes and edges. Hence, the goal is to strike a balance that ensures efficiency without sacrificing the incorporation of essential graph components.

\paragraph{Node selection} In this step, the aim is to select relevant text spans to serve as nodes in the initial graph. For each span $(i,j)$ within the span set $\mathcal{S}$, the model computes a score:
\begin{equation}
    \texttt{sel\_node}((i,j)) = \sigma(\vw_{n} ^T \vs_{ij})
    \label{eq:nodesel}
\end{equation}

where $\vw_{n} \in \mathbb{R}^{D\times 1}$ is a learned weight matrix, and $\sigma$ the sigmoid function. Finally, the top-$K$ spans that have the highest $\texttt{sel\_node}$ score are selected as nodes. We denote these nodes as $\mathcal{V}$.

\paragraph{Edge selection} This layer prunes the edges of the fully connected graph formed by all the nodes in $\mathcal{V}$. It first computes the score of each potential directed edge $(i,j) \xrightarrow{} (k,l)$ — with both source and target nodes belonging to the previously computed $\mathcal{V}$:
\begin{equation}
    \texttt{sel\_edge}((i,j), (k,l)) = \vw_{e} ^T [\vs_{ij};\vs_{kl}]
    \label{eq:edgesel}
\end{equation}

where $\vw_{e} \in \mathbb{R}^{2D\times 1}$ is a learned weight matrix. The top-$K$ edges with the highest scores are then selected; we denote them as $\mathcal{E}$.

\paragraph{Initial graph formation} 
After the node $\mathcal{V}$ and edge $\mathcal{E}$ selections, we construct the initial graph, denoted as $G = (\mathcal{V}, \mathcal{E})$, illustrated in step 4 of Figure \ref{fig:architecture}. This graph forms the basis for the subsequent structure learning phase, whose objective is to refine and edit $\mathcal{G}$ into the final IE graph.

\section{Structure Learning}

The goal of structure learning is to modify the structure of the graph constructed previously to produce the final IE graph. This process unfolds in two main steps: Utilizing a Graph Neural Network (GNN), the structure learner first enriches the representations of nodes and edges in $\mathcal{G}$, leveraging the existing graph's information. Then, the structure learner performs editing operations on $\mathcal{G}$, using the learned structure-aware representations. 

\subsection{Graph Representation Learning \label{trans_tok}} This layer aims to enrich the representations of nodes and edges using information from the previously constructed graph $\mathcal{G}$. Typically, graph representation learning employs message-passing Graph Neural Networks (MPGNNs), which aggregate information from neighboring nodes. However, we find this approach yields suboptimal performance due to the noisy and heterogeneous nature of our input graph. Moreover, standard message-passing GNNs often encounter challenges such as oversmoothing \citep{chen2019measuring} and oversquashing \citep{alon2021bottleneck}. Given these limitations, we use TokenGT \citep{kim2022pure}, which treats nodes and edges of the graph as independent tokens, and feeds these tokens as input to a standard Transformer \citep{NIPS2017_3f5ee243}. TokenGT has been proven to be more expressive than message-passing GNNs and provides stronger empirical performance. It effectively addresses the shortcomings of MPGNNs and facilitates long-range interactions thanks to global attention mechanisms.

\paragraph{Graph Tokenization} Let's consider two nodes $n=(i,j)$ and $m=(k,l)$, representing spans, and a directed edge $n \xrightarrow{} m$ in graph $\mathcal{G}$. The representation of the nodes $n$ and $m$ are computed as:
\begin{equation}
\begin{split}
    \vz_{n} = \vs_{n} + [\vp_{n};\vp_{n}] &\\
    \vz_{m} = \vs_{m} + [\vp_{m}; \vp_{m}] &
\end{split}
\end{equation}
and the representation of the edge $n \xrightarrow{} m$ is:
\begin{align}
\vz_{n,m} &= [\vp_{n}; \vp_{m}]
\label{eq:edge}
\end{align}


Where $\mathbf{s}_{n}$ and $\mathbf{s}_{m}$ ($\in \mathbb{R}^D$) are the representations of the spans (as computed in Equation \ref{eq1}), and $\mathbf{p}_n$ and $\mathbf{p}_m$ ($\in \mathbb{R}^{D/2}$) correspond to node identifiers. The node identifiers are initialised with orthonormal vectors (following \citep{kim2022pure}) and then updated during training. During forward passes, they are randomly assigned to each node of the graph. We found that freezing the node identifiers can also work but provides suboptimal results. Furthermore, as noted in Equation \ref{eq:edge}, the edge representation consists solely of node identifiers. By doing so (not providing edge-specific features), we ensure the transformer layer purely uses the graph structure to represent the edges, which it does surprisingly well. However, explicitly integrating edge features can sometimes enhance the results.

\paragraph{Transformer layer} The transformer layer takes as input the node and edge tokens computed previously. Following \citep{kim2022pure}, we augment the input with learned token type embeddings $\ve_{Node}$ and $\ve_{Edge}$ ($\in \mathbb{R}^D$), which specifies whether a token represents a node or an edge. Then, we linearly project the token embeddings to match the dimensions of the transformer layer:
\begin{equation}
\begin{split}
    \vz_n^{(0)} &= \mW_{in} ^T (\vz_n + \ve_{Node}) \\
    \vz_{n,m}^{(0)} &= \mW_{in} ^T (\vz_{n,m} + \ve_{Edge})
\end{split}
\end{equation}

where, $\mW_{in} \in \mathbb{R}^{D \times D}$ is the input projection weight. Then, the stacked nodes and edges, denoted as $\mZ^{(0)} \in \mathbb{R}^{(|\mathcal{V}|+|\mathcal{E}|) \times D}$ are fed into an $L$-layer transformer to produce the final representation $\mZ^{(L)}$.

\subsection{Graph editing}
In this stage, the goal is to obtain the final graph structure (IE graph), comprising nodes and their connectivity. This is accomplished using the representations $\mZ^{(L)}$ learned by the graph transformer in the preceding subsection. We use a similar approach to the Graph Edit Network (GEN) \citep{paassen2021graph}. Following GEN, our layer either keeps or removes elements from the graph. However, in contrast to GEN, we do not introduce new elements, assuming that the initial graph $\mathcal{G}$ already contains all necessary nodes and edges. Consequently, since no new nodes are added, graph editing can be executed in a single stage.

\paragraph{Edit layer} 
The graph editing process computes the probability for each node ($p_{\text{keep}}(n)$) and edge ($p_{\text{keep}}(n, m)$) regarding whether they should be kept or removed using a linear layer:
\begin{equation}
\begin{split}
    p_{\texttt{keep}}(n) &= \sigma(\vw_k ^T \vz_n^{(L)}) \\
    p_{\texttt{keep}}(n, m) &= \sigma(\vw_k ^T \vz_{n,m}^{(L)})
\end{split}
\end{equation}

where $\vw_k \in \mathbb{R}^D$ is a learnable weight shared by the edges and the nodes.

\paragraph{Final graph structure} To predict the final graph structure, we select the final nodes $\mathcal{V}_{f}$ and final edges $\mathcal{E}_{f}$, according to their \textit{keep} probability, as follows:
\begin{equation}
\begin{split}
    \mathcal{V}_{f} =  \left\{ n \in \mathcal{V}: p_{\texttt{keep}}(n) > \texttt{0.5} \right\} &\\
    \mathcal{E}_{f} = \left\{(n \xrightarrow{} m) \in \mathcal{E}: p_{\texttt{keep}}(n,m) > \texttt{0.5}\right\} &
\end{split}
\end{equation}

Depending on the specific text-graph dataset used, it might be necessary to ensure that the final nodes $\mathcal{V}_{f}$ do not have overlapping spans. To address this, we employ a greedy algorithm that iteratively selects the highest-scoring node while respecting this constraint \citep{zaratiana-etal-2022-global,zaratiana-etal-2022-named}. Additionally, although rare, there may be instances where an edge $(n \rightarrow m)$ is selected in $\mathcal{E}_{f}$, while either $n$ or $m$ is not in $\mathcal{V}_{f}$. In such cases, we simply discard the edge to maintain consistency in the graph structure.

\subsection{Classification}

The final step for obtaining the IE graph involves labeling the nodes and edges of the graph. We do so by computing classification scores for nodes and edges, denoted as $y_n$ and $y_{n,m}$ respectively, using two independent feed-forward networks:
\begin{equation}
\begin{split}
    \vy_n &= \text{FFN}_{\mathcal{V}}(\vz_{n}^{(L)}) \in \mathbb{R}^{|\mathcal{C}|} \\
    \vy_{n,m} &= \text{FFN}_{\mathcal{E}}(\vz_{n,m}^{(L)}) \in \mathbb{R}^{|\mathcal{R}|}
\end{split}
\end{equation}

where $|\mathcal{C}|$ is the number of node types, and $|\mathcal{R}|$ is the number of entity types. 


\section{Training}
\textit{GraphER} is trained using multitask learning \citep{Caruana1997MultitaskL}, by jointly optimising the loss for the different components, treated as \textit{independent classifiers} \citep{Punyakanok2005LearningAI}. The total loss function, \(\mathcal{L}_{total}\), for training our model is computed as:
\begin{equation}
    \mathcal{L}_{total} = \mathcal{L}_{\mathcal{V}} + \mathcal{L}_{\mathcal{E}} + \mathcal{L}_{edit} + \mathcal{L}_{cls}
\end{equation}

This equation sums up the node selection loss (\(\mathcal{L}_{\mathcal{V}}\)), edge selection loss (\(\mathcal{L}_{\mathcal{E}}\)), edit losses (\(\mathcal{L}_{edit}\)), and final node/edge classification losses (\(\mathcal{L}_{cls}\)). Detailed explanations of each component are presented in Appendix \ref{loss:det}.

\section{Experimental setup}
\subsection{Datasets}
We evaluate \textit{GraphER} on three datasets for joint entity-relation extraction, namely SciERC \citep{luan-etal-2018-multi}, CoNLL04 \citep{carreras-marquez-2004-introduction}, and ACE 05 \citep{ace05}. We describe them in the following and provide details and statistics about the datasets in Table \ref{tab:dataset_statistics}.

\paragraph{ACE 05} was collected from a variety of domains, such as newswires, online forums, and broadcast news. It provides a very diverse set of entity types, such as Persons (``\textit{PER}''), Locations (``\textit{LOC}''), Geopolitical Entities (``\textit{GPE}''), and Organizations (``\textit{ORG}''), as well as complex types of relationships between them, including General Affiliations (``\textit{GEN-AFF}''), Personal Social Relationships (``\textit{PER-SOC}''), among others. This dataset is particularly notable for its complexity and wide coverage of entity and relation types, making it a robust benchmark for evaluating the performance of joint information extraction models.

\paragraph{CoNLL-2004} is a popular benchmark dataset for entity-relation extraction in English. It focuses on general entities, such as People, Organizations, and Locations, and simple relation types, such as ``\textit{Work\_For}''  and ``\textit{Live\_In}''.

\paragraph{SciERC} is specifically designed for the AI domain. It includes entity and relation annotations from a collection of documents from 500 AI paper abstracts. It contains scientific entity types and relation types and is primarily intended for scientific knowledge graph construction.

\subsection{Evaluation Metrics}

For the named entity recognition (NER) task, we use span-level evaluation, demanding precise entity boundary and type predictions. For evaluating relations, we employ two metrics: (1) Boundary Evaluation (\textbf{REL}), which requires correct prediction of entity boundaries and relation types, and (2) Strict Evaluation (\textbf{REL+}), which also necessitates correct entity type prediction. We report the micro-averaged F1 score following previous works.

\begin{table}[]
\centering
\resizebox{\columnwidth}{!}{%
\begin{tabular}{@{}lcccccc@{}}
\toprule
Dataset & \(|\mathcal{E}|\) & \(|\mathcal{R}|\) & \# Train & \# Dev & \# Test \\
\midrule
ACE05 & 7 & 6 & 10,051 & 2,424 & 2,050 \\
CoNLL 2004 & 4 & 5 & 922 & 231 & 288 \\
SciERC & 6 & 7 & 1,861 & 275 & 551 \\
\bottomrule
\end{tabular}%
}
\caption{Number of entity labels, relation labels, and sentences in each dataset.}
\label{tab:dataset_statistics}
\end{table}

\begin{table}[]
\centering
\label{tab:hyperparameters}
\resizebox{\columnwidth}{!}{%
\begin{tabular}{@{}lccc@{}}
\toprule
Hyperparameter & ACE 05 & CoNLL 2004 & SciERC \\ 
\midrule
Backbone      & ALB    &  ALB  & SciB  \\
GNN Layers      & 2    &  2  & 2  \\
Optimizer      & AdamW    &  AdamW  & AdamW  \\
lr backbone     & 1e-5  &    3e-5        &   3e-5     \\
lr others         &   5e-5     &   5e-5         & 5e-5       \\
Weight Decay   &   1e-2     &  1e-4          &  1e-4      \\
FNN Dropout    &  0.1      &   0.1         &   0.1     \\
Hidden Size    &  768      &  768          &    768    \\
Training Steps &  120k     &  50k          &   30k \\ Warmup         &  5000      &   5000         &  3000      \\Batch size &  8     &  8          &   8    \\ Span length &  12     &  12          &   12   \\ 
\bottomrule
\end{tabular}%
}
\caption{\textbf{Hyperparameters Recapitulation}. ALB denotes \texttt{albert-xxlarge-v1} and SciB denotes \texttt{scibert\_scivocab\_uncased}.}
\label{tab:hyper}
\end{table}

\subsection{Hyperparameters}
In this study, we implemented \textit{GraphER} using ALBERT \citep{Lan2019ALBERTAL} for the ACE 05 and CoNLL 2004 datasets, and SciBERT \citep{beltagy-etal-2019-scibert} for the SciERC dataset, aligning with previous works. For all the models, we used the AdamW optimizer, and the learning rates and the number of training steps were tuned differently according to the dataset. We used two layers for the transformer layer, using the standard variant from \citet{NIPS2017_3f5ee243} for structure learning. For the top-K value used for node and edge selection, we set it to the same length as the input sequence, which proved satisfactory in preliminary experiments. We detail the hyperparameters in Table \ref{tab:hyper}. \textit{GraphER} was implemented using PyTorch and trained on a server equipped with A100 GPUs.

\begin{table*}[]
\renewcommand{\arraystretch}{1.3} 
\centering

\resizebox{\textwidth}{!}{%
\begin{tabular}{l|l|ccc|ccc|ccc}
\toprule
& & \multicolumn{3}{c|}{\textbf{ACE 05}} & \multicolumn{3}{c|}{\textbf{CoNLL 2004}} & \multicolumn{3}{c}{\textbf{SciERC}} \\
\cmidrule(lr){3-5} \cmidrule(lr){6-8} \cmidrule(lr){9-11} 
Model & Backbone & Entity & REL & REL+ & Entity & REL & REL+ & Entity & REL & REL+ \\
\midrule
DYGIE++ \citep{wadden-etal-2019-entity} & BB \& SciB & 88.6 & 63.4 & -- & -- & -- & -- & 67.5 & \underline{48.4} & -- \\
Tab-Seq \citep{wang-lu-2020-two} & ALB & 89.5 & -- & 64.3 & 90.1 & 73.8 & 73.6 & -- & -- & -- \\
PURE \citep{zhong-chen-2021-frustratingly} & ALB \& SciB & 89.7 & 69.0 & 65.6 & -- & -- & -- & 66.6 & 48.2 & 35.6 \\
PFN \citep{yan-etal-2021-partition} & ALB \& SciB & 89.0 & -- & 66.8 & -- & -- & -- & 66.8 & -- & 38.4 \\
UniRE \citep{wang-etal-2021-unire} & ALB \& SciB & 89.9 & -- & 66.0 & -- & -- & -- & \underline{68.4} & -- & 36.9 \\
TablERT \citep{ma-etal-2022-joint} & ALB & 87.8 & 65.0 & 61.8 & \textbf{90.5} & 73.2 & 72.2 & -- & -- & -- \\
UTC-IE \citep{yan-etal-2023-utc} & ALB \& SciB & 89.9 & -- & \textbf{67.8} & -- & -- & -- & 69.0 & -- & 38.8 \\ 
\midrule
\textbf{\textit{GraphER}} & ALB \& SciB & 89.8 & 68.4 & 66.4 & 89.6 & \underline{76.5} & \underline{76.5} & \textbf{69.2} & \textbf{50.6} & \underline{39.1} \\
+ \textit{Edge features} & ALB \& SciB & 89.8 & 68.0 & 66.0 & \underline{90.2} & \textbf{76.6} & \textbf{76.6} & 68.0 & 50.4 & \textbf{39.4} \\
\bottomrule
\end{tabular}%
}
\caption{\textbf{Results for different approaches and datasets}. "Entity" refers to the F1 score for entity recognition, "REL" for relaxed relation extraction, and "REL+" for strict relation extraction. The "Backbone" column indicates the underlying architecture for each model (ALB for \texttt{albert-xxlarge-v1}, BB for \texttt{bert-base-cased}, and SciB for \texttt{scibert-base-uncased}).}
\label{tab:comparison}
\end{table*}

\subsection{Baselines} We primarily compare \textit{GraphER}, with comparable approaches from the literature in terms of model size. \textbf{DyGIE++} \citep{wadden-etal-2019-entity} is a model that uses a pretrained transformer to compute contextualised representations and enriches the representations of spans using graph propagation.\textbf{PURE} \citep{zhong-chen-2021-frustratingly} is a pipeline model for the information extraction task that learns distinct contextual representations for entities and relations. \textbf{PFN} \citep{yan-etal-2021-partition} introduces methods that model two-way interactions between the task by partitioning and filtering features. \textbf{UniRE} \citep{wang-etal-2021-unire} proposes a joint entity and relation extraction model that uses a unified label space for entity and relation classification. \textbf{Tab-Seq} \citep{wang-lu-2020-two} tackles the task of joint information extraction by treating it as a table-filling problem. Similarly, in \textbf{TablERT} \citep{ma-etal-2022-joint}, entities and relations are treated as tables, and the model utilizes two-dimensional CNNs to effectively capture and model local dependencies within these table-like structures. Finally, \textbf{UTC-IE} \citep{yan-etal-2023-utc} treats the task as token-pair classification. It incorporates Plusformer to facilitate axis-aware interactions through plus-shaped self-attention and local interactions using CNNs over token pairs.

\section{Results and Analysis}
\subsection{Main results} The main results of our experiments are reported in Table \ref{tab:comparison}. For \textit{GraphER}, we report two variants: with and without edge features. Interestingly, there is only a marginal distinction between these variants, suggesting that utilizing node identifiers alone adequately represents edges for the graph transformer layer. Compared to state-of-the-art baselines, our proposed model achieves the highest performance on both CoNLL 2004 and SciERC datasets. Notably, \textit{GraphER} outperforms the best-performing approach on CoNLL by more than 3 points in relation F1. Finally, while \textit{GraphER}'s performance on ACE 05 exhibits slightly lower results in relation evaluation, its entity evaluation performance matches state-of-the-art results.

\begin{table}[h]
    \centering
    {\fontsize{10pt}{12pt}\selectfont
    \begin{tabular}{llcccc}
        \toprule
        \textbf{Dataset} & \textbf{Setting} & \textbf{ENT} & \textbf{REL} & \textbf{REL+} \\
        \midrule
        \textbf{ACE 05} & \texttt{Trans}   & \textbf{89.8} & \textbf{68.4} & \textbf{66.4} \\
                   & \texttt{GCN}  & 88.3 & 55.3 & 52.1 \\
               & \texttt{GAT}  & 88.3 & 54.5 & 52.1 \\
               & \texttt{SAGE} & 57.2 & 56.8 & 53.2 \\
        \midrule
        \textbf{CoNLL 04} & \texttt{Trans}   & 89.6 & \textbf{76.5} & \textbf{76.5} \\
                   & \texttt{GCN}  & \textbf{89.7} & 72.4 & 72.4 \\
               & \texttt{GAT}  & 86.1 & 70.6 & 70.6 \\
               & \texttt{SAGE} & 88.8 & 72.2 & 72.2 \\
        \midrule
        \textbf{SciERC} & \texttt{Trans}   & \textbf{69.2} & \textbf{50.6} & \textbf{39.1} \\
                   & \texttt{GCN}  & 49.8 & 32.7 & 17.5 \\
               & \texttt{GAT}  & 30.9 & 31.6 & 15.0 \\
               & \texttt{SAGE} & 36.2 & 37.1 & 19.3 \\
        \bottomrule
    \end{tabular}
    }
    \caption{Comparison of graph transformers against message-passing based GNN: Graph Convolution Network (GCN), Graph Attention Network (GAT) and GraphSAGE (SAGE).}
\label{abl:attention}
\end{table}

\subsection{Comparison with Message Passing GNN}
In our study, we primarily utilize a transformer to learn the structure of the information extraction graph. In this section, we contrast it with the traditional Message Passing Graph Neural Network (MPGNN), where each node in the graph can only communicate messages to its immediate neighboring nodes for each layer. To achieve this, we substitute the transformer layer with MPGNN to learn the representation of the spans, i.e., nodes. The numerical results of this comparative analysis are presented in Table \ref{abl:attention}.

\paragraph{Message Passing} 
First, let's briefly outline the concept of message passing. During each iteration of message passing in a GNN, a hidden embedding $\vz_n$ corresponding to each node $n$ is updated based on information aggregated from $n$'s incoming nodes (${m | m \rightarrow n \in \mathcal{E}})$ in the graph, denoted as $\mathcal{N}(n)$. A message passing update can be expressed as follows:
\begin{equation}
    \vz_n^{(k+1)} = \psi \left(\vz_n^{(k)}, \bigoplus_{m \in \mathcal{N}(n)} \phi(\vz_n^{(k)}, \vz_m^{(k)}) \right) 
    \label{equ:mess}
\end{equation}

\noindent Where $\psi$ and $\phi$ are learnable functions, and $\bigoplus_{m}$ represents a permutation-invariant operation such as summation or max pooling. In our study, we explore three variants of the message-passing GNN: Graph Convolutional Network (GCN) \citep{kipf2018neural}, Graph Attention Network (GAT) \citep{veličković2018graph}, and GraphSAGE (SAGE) \citep{Hamilton2017InductiveRL}. The only distinction among these variants lies in the aggregation function $\phi$ for message passing and the permutation-invariant operation $\bigoplus$ in Equation \ref{equ:mess}. For all variants, we utilized the available implementations in the PyTorch Geometric library \citep{fey2019fast}.

\begin{figure}
    \centering
\includegraphics[width=1.\columnwidth]{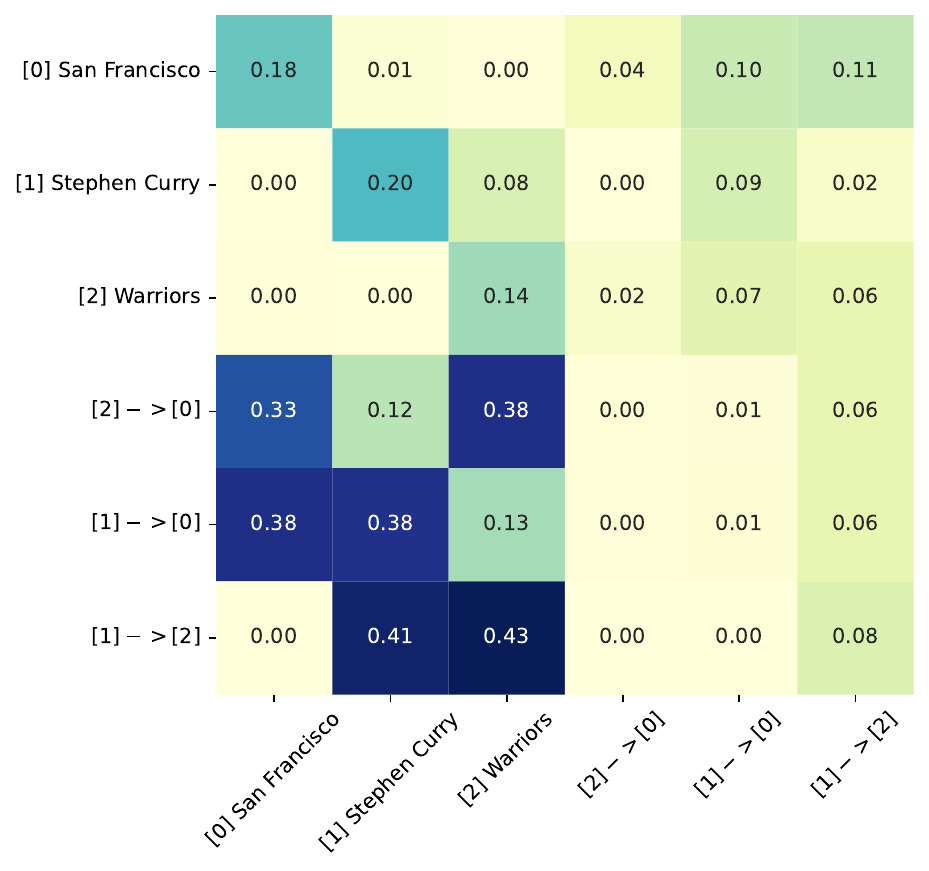}
        \caption{\textbf{Attention visualization in Graph Transformer}. In this figure, we illustrate the attention for the input ``\textit{\underline{Stephen Curry} plays for the \underline{Warriors} in \underline{San Francisco}}''. The nodes are represented by their identifier and text spans and edges are represented by their corresponding (directed) pairs of node identifier. We illustrate only the top 3 nodes and top 3 edge for better visibility.}
    \label{fig:attviz}
\end{figure}

\paragraph{Results} The results for each variant are reported in Table \ref{abl:attention}. The transformer variant consistently outperforms the MPGNN baselines by a significant margin in terms of relation evaluation. For instance, in terms of \textit{REL+} metric, it outperforms the variants by more than 10 points on ACE 05, more than 4 points on CoNLL 04, and over 20 points on SciERC. The results of MPGNN on CoNLL 04 are relatively more competitive compared to other datasets, possibly due to its simplicity (generic entity and relation types), requiring less reasoning for the model. Our main intuition for the poor performance of the MPGNN approach is as follows: (1) Our constructed graph is highly noisy at both node and edge levels. The functionality of MPGNN, which forces neighborhood direct connections, makes it difficult to distinguish noise from valid signals. In contrast, the transformer layer allows all nodes and edges to have a global overview of the entire graph, facilitating the discrimination between signal and noise. (2) Additionally, the information extraction graph is highly heterogeneous (many node and edge types), while MPGNNs are more suited for homogeneous and homophilic graphs. (3) Finally, MPGNNs are prone to problems such as oversmoothing and oversquashing, which can result in suboptimal learned representations. Transformers are less prone to these problems.

\subsection{Attention Analysis} In Figure \ref{fig:attviz}, we illustrate the attention map generated by our graph transformers. The resulting map exhibits intuitive patterns: first, we observe that nodes assign high attention weight to themselves, indicating that a significant part of the information is already encoded within the span representation. In addition, spans also show considerable attention to edges, indicating that the graph structure may be useful to improve node representation to some extent. Furthermore, we also observe interesting patterns concerning relations. Notably, relations that are solely based on node identifiers can precisely attend to their corresponding head and tail nodes. For instance, the edge ``$[2]->[0]$'' assigns the highest attention to the nodes with identifier [2] and [0], which corresponds to ``\textit{Warriors}'' and ``\textit{San Francisco}'' respectively. This observation applies to the other edges. This suggests that by solely using identifiers to represent edges, the transformer layer effectively reconstructs the graph structure.

\begin{figure}[h]
    \centering
\includegraphics[width=1.\columnwidth]{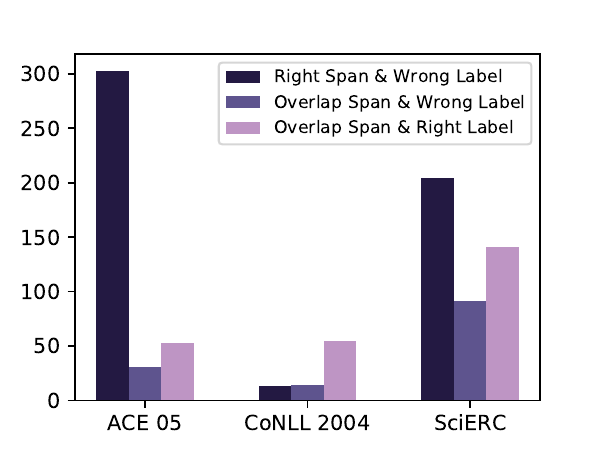}
\caption{\textbf{Common Errors in Entity Recognition}. This caption highlights prevalent model errors found within false positives.}
    \label{fig:entity_errors}
\end{figure}

\section{Error Analysis}

\begin{figure*}[ht]
    \centering
\includegraphics[width=1.\textwidth]{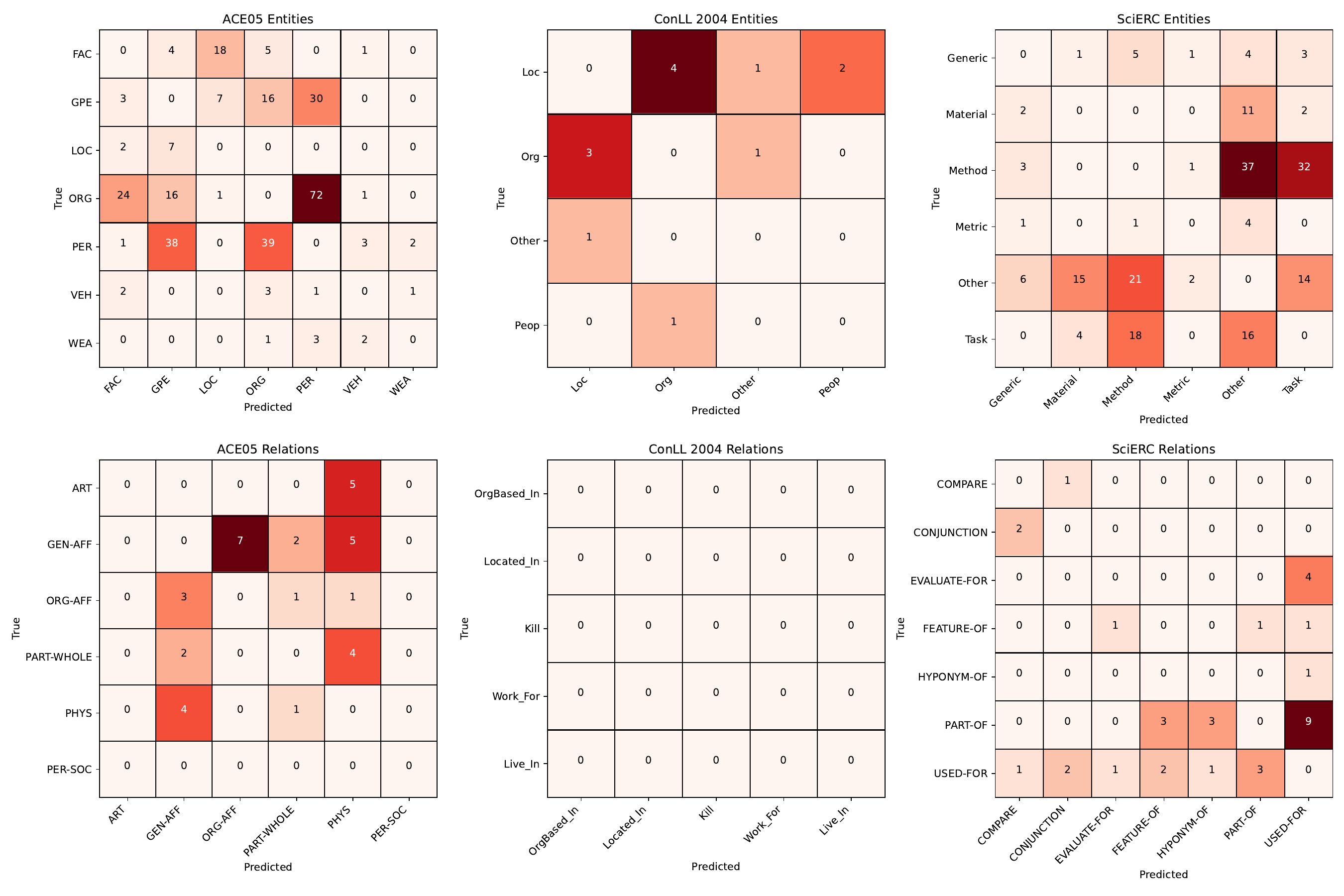}
        \caption{\textbf{Entity and Relation confusion matrix.}}
    \label{fig:all_misclassifications}
\end{figure*}

We conduct a detailed error analysis on the test sets, evaluating \textit{GraphER}'s performance and pinpointing improvement areas (see Figure \ref{fig:all_misclassifications}).


\paragraph{Entity errors} Our analysis of entity extraction errors is illustrated in Figure \ref{fig:entity_errors}. For ACE 05, errors primarily come from correct span identification but incorrect label prediction. Notably, a significant portion of these errors involves pronominal entities like \textit{we}, \textit{us}, \textit{it}, and \textit{our}. In contrast, we observe few labeling errors with the CoNLL 2004 dataset, due to its simple entity types and predominantly exact span identification. However, in SciERC, an important number of misclassifications occur, primarily attributed to underspecified relation types such as ``\textit{Generic}'' and ``\textit{OtherScientificTerm}'', as well as the confusion between ``\textit{Method}'' and ``\textit{Task}'' entity types.

\paragraph{Relation errors}
For relations, we concentrate on analysing labeling errors, assuming spans are predicted correctly. In the ACE 05 dataset, errors involving \texttt{GEN-AFF} represent 40\% of errors, often confused with \texttt{ORG-AFF} due to their semantic similarities, which can even challenge human discernment. In SciERC, 70\% of labeling errors involve \texttt{PART-OF} and \texttt{USED-FOR} types, which are often confused with each other. Finally, for the CoNLL 2003 dataset, there are no labeling errors for the relations; the errors are mainly false positives and false negatives. This highlights that this dataset is less challenging, but also that our model is strong.

\section{Related Works}

\paragraph{Joint IE} 
The field of information extraction (IE) has evolved from traditional pipeline models, which handle entity recognition \citep{Chiu2015NamedER} and relation extraction \citep{Zelenko2002KernelMF, Bach2007ARO, Lin2016NeuralRE} sequentially, to end-to-end models. These approaches aim to mitigate error propagation \citep{10.1007/10704656_11, Nadeau2007ASO} by jointly optimizing entity and relation extraction \citep{roth-yih-2004-linear, Sun_Zhang_Mensah_Mao_Liu_2021}, enhancing the interaction between the two tasks and overall performance. Proposed approaches include table-filling methods \citep{wang-lu-2020-two, ma-etal-2022-joint}, span pair classification \citep{Eberts2019SpanbasedJE, wadden-etal-2019-entity, ye-etal-2022-packed}, set prediction \citep{Sui2020JointEA}, augmented sequence tagging \citep{ji-etal-2020-span}, and the use of unified labels for the task \citep{wang-etal-2021-unire, yan-etal-2023-utc}. Additionally, the usage of generative models \citep{Achiam2023GPT4TR} has become popular for this task, where input texts are encoded and decoded into augmented language \citep{paolini2021structured}. Some of these approaches conduct fine-tuning on labeled datasets \citep{lu-etal-2022-unified, fei2022lasuie,Zaratiana_Tomeh_Holat_Charnois_2024}, while others use prompting techniques on large language models such as ChatGPT \citep{wadhwa-etal-2023-revisiting}. Diverging from these approaches, our proposed model tackles the joint IE task as Graph Structure Learning, where the structure of information extraction is first inferred, followed by the prediction of node and edge types to more effectively incorporate structural information.

\paragraph{GSL} 
Graph Structure Learning is a crucial task aimed at jointly optimizing the graph structure and downstream performance during training \citep{zhu2022survey,zhou2023opengsl}. In this context, graphs may be incomplete, and in some cases, missing entirely, as they can be in our study. Different models employ various approaches to infer edges between nodes. For example, a metric learning approach utilizes a metric function on pairwise node embeddings to derive edge weights \citep{li2018adaptive,yu2020graphrevised,zhang2020gnnguard}. Others utilize more expressive neural networks to infer edge weights based on node representations \citep{pmlr-v119-zheng20d,10.1145/3437963.3441734,veličković2020pointer,sun2021graph}. Alternatively, they may treat the adjacency matrix as learnable parameters and directly optimize them along with GNN parameters \citep{gao2019exploring,jin2020graph}. In our paper, we approach joint entity and relation extraction as Graph Structure Learning (GSL) by first inferring the structure of the graph, where text spans are nodes and relations are edges, and then predicting the types of its elements.

\section{Conclusion} In this work, we approached the task of joint entity and relation extraction as Graph Structure Learning. Our methodology involves predicting an initial and noisy graph, followed by leveraging cutting-edge techniques from the Graph Neural Network (GNN) literature to refine the graph representations. This refinement process results in the production of the final information extraction graph through graph editing. When evaluated on common IE benchmarks, \textit{GraphER} demonstrates strong performance compared to state-of-the-art approaches.

\section*{Limitations}
While \textit{GraphER} demonstrates strong performance on the SciERC and CoNLL 04 datasets, it exhibits limitations when applied to the ACE 05 dataset. Particularly, it tends to make errors when entities are ambiguous, especially in cases where the span alone does not provide sufficient context to infer the type of entities. This is particularly evident with pronouns, which require additional contextual information for accurate classification. We hypothesize that this limitation may stem from suboptimal and simplistic span representation (concatenation), resulting in the loss of critical contextual information. Further exploration and refinement of span representation techniques may be necessary to address this issue effectively.

\section*{Acknowledgments}  This work was granted access to the HPC resources of IDRIS under the allocation 2023-AD011014472 and AD011013682R1 made by GENCI. This work is partially supported by a public grantoverseen by the French National Research Agency
(ANR) as part of the program Investissements
d’Avenir (ANR-10-LABX-0083).

\bibliography{custom,anthology}

\appendix

\section{Training details \label{loss:det}}
\subsection{Notations}
We define $l_n$ and $l_{n, m}$ as the gold-standard labels for nodes (spans) and edges in the true graph $\mathcal{G}_g = (\mathcal{V}_g, \mathcal{E}_g)$. For spans not in $\mathcal{V}_g$ (i.e., $n \not \in \mathcal{V}_g$), we assign $l_n = \texttt{non-entity}$. Similarly, for edges not in $\mathcal{E}_g$ (i.e., $n \xrightarrow{} m \not \in \mathcal{E}_g$), we assign $l_{n, m} = \texttt{no-relation}$. We define $\delta_n$ and $\delta_{n,m}$ as indicators: $\delta_n = 0$ if $l_n = \texttt{non-entity}$ and $\delta_{n,m} = 0$ if $l_{n,m} = \texttt{no-relation}$; they are $1$ otherwise. The binary cross-entropy loss $\texttt{BCE}$ is defined as:
\begin{equation}
    \texttt{BCE}(p, \hat{p}) = -\left(p \log(\hat{p}) + (1 - p) \log(1 - \hat{p})\right)
\end{equation}
where $y$ is the true (binary) label and $\hat{y}$ is the predicted probability. The categorical cross-entropy loss $\texttt{CE}$, which computes the negative log-probability of the true classes, is defined as:
\begin{equation}
\texttt{CE}(l, \vy) = - \log(\texttt{softmax}(\vy)[l])
\end{equation}
where $l$ is the true label index, and $\vy \in \mathbb{R}^{|\mathcal{L}|}$ is the unnormalised logit vector, with $\mathcal{L}$ representing the set of possible classes. The softmax function, denoted as $\texttt{softmax}(\vy)$, converts the logit vector $\vy$ into a probability distribution over the classes. It is defined as:
\begin{equation}
\texttt{softmax}(\vy)[i] = \frac{\exp(\vy_i)}{\sum_{j=1}^{|\mathcal{L}|} \exp(\vy_j)}
\end{equation}
for each component $i$ of the vector $\vy$. This function ensures that the output probabilities sum to 1 and are normalised across all classes.

\subsection{Loss functions}
 \paragraph{Node selection loss} The node selection loss seek to increase the score of nodes $n$ that are contained in $\mathcal{V}_g$:

 \begin{equation}
     \mathcal{L}_{\mathcal{V}} = \sum_{n \in \mathcal{S}} \texttt{BCE}(\delta_{n}, q_{n})
 \end{equation} 

 where, $q_n = \texttt{sel\_node(n)}$ computed in equation \ref{eq:nodesel}, and $\mathcal{S}$ is a set containing all possible span (up to a maximum length) in the input text.

\paragraph{Edge selection loss} The edge selection loss seeks to increase the score of edges $n \xrightarrow{} m$ that are contained in $\mathcal{E}_g$:

 \begin{equation}
     \mathcal{L}_{\mathcal{E}} = \sum_{(n,m) \in \mathcal{V}^2} \texttt{BCE}(\delta_{n,m}, q_{n,m})
 \end{equation} 


  where, $q_{n,m} = \texttt{sel\_edge(n,m)}$, the edge selection score, computed in Equation \ref{eq:edgesel}.

\paragraph{Edit loss} The edit loss seeks to maximize the probability of keeping gold nodes and edges, after refining their representation using a graph transformer:
 \begin{equation}
 \begin{split}
     \mathcal{L}_{edit} = & \sum_{n \in \mathcal{V}} \texttt{BCE}(\delta_{n}, p_{\texttt{keep}}(n)) \\ + & \sum_{(n, m) \in \mathcal{E}} \texttt{BCE}(\delta_{n,m}, p_{\texttt{keep}}(n,m))
 \end{split}
 \end{equation} 

 \noindent This corresponds to the sum of node-level and edge-level edit loss, over the graph $\mathcal{G}$ with nodes $\mathcal{V}$ and edges $\mathcal{E}$.

\paragraph{Classification loss} The classification losses optimize the gold label nodes and edges of the final graph $\mathcal{G}_f$:
 \begin{equation}
 \begin{split}
     \mathcal{L}_{cls} = & \sum_{n \in \mathcal{V}_f} \texttt{CE}(l_{n}, \vy_{n}) + \sum_{(n, m) \in \mathcal{E}_f} \texttt{CE}(l_{n,m}, \vy_{n,m})
 \end{split}
 \end{equation} 

 Finally, during training, our model minimizes the sum of all losses. For simplicity, we do not weigh the different loss contributions as usually done in multitask learning.

\end{document}